# Adapting an Artificial Intelligence Sexually Transmitted Diseases Symptom Checker Tool for Mpox Detection: The HeHealth Experience

Rayner Kay Jin Tan[1,2], Dilruk Perera[1,2], Salomi Arasaratnam[2], Yudara Kularathne[2]

[1]Saw Swee Hock School of Public Health, National University of Singapore and National University Health System, Singapore, Singapore

[2]HeHealth.ai, Singapore, Singapore

Corresponding Author: Dr Rayner Kay Jin Tan

12 Science Drive 2, MD1 Tahir Foundation Building #10-01

Singapore 117549

Phone: +6591878576

Email: Rayner.tan@nus.edu.sg

**Abstract**

Artificial Intelligence (AI) applications have shown promise in the management of pandemics and have been widely used to assist the identification, classification, and diagnosis of medical images. In response to the global outbreak of Monkeypox (Mpox), the HeHealth.ai team leveraged an existing tool to screen for sexually transmitted diseases (STD) to develop a digital screening test for symptomatic Mpox through AI approaches. Prior to the global outbreak of Mpox, the team developed a smartphone app (HeHealth) where app users can use their own smartphone cameras to take pictures of their own penises to screen for symptomatic STD. The AI model was initially developed using 5000 cases and use a modified convolutional neural network (CNN) to output prediction scores across visually diagnosable penis pathologies including Syphilis, Herpes Simplex Virus (HSV), and Human Papilloma Virus (HPV). From June 2022 to October 2022, a total of about 22,000 users had downloaded the HeHealth app, and about 21,000 images have been analysed using HeHealth AI technology. We then engaged in formative research, stakeholder engagement, rapid consolidation images, a validation study, and implementation of the tool from July 2022. From July 2022 to October 2022, a total of 1000 Mpox-related images had been used to train the Mpox symptom checker tool. Our digital symptom checker tool showed accuracy of 87% to rule in Mpox and 90% to rule out symptomatic Mpox. Several hurdles identified included issues of data privacy and security for app users, initial lack of data to train the AI tool, and the potential generalizability of input data. We offer several suggestions to help others get started on similar projects in emergency situations, including engaging a wide range of stakeholders, having a multidisciplinary team, prioritizing pragmatism, as well as the concept that 'big data' in fact is made up of 'small data'.

**Introduction**

Monkeypox (Mpox) was declared a public health emergency of international concern (PHEIC) on 23$^{rd}$ July 2022, owing to the risk of human-to-human transmission, and the rapidly increasing number of cases being recorded worldwide.[1] An important observation of the ongoing global outbreak of Mpox is that the majority of the infected cases are among men who have sex with men, and this has led to stigmatizing attitudes that conflate both Mpox acquisition with one's sexual identity. This has led to concerns that individuals may be disincentivized to seek care due to concerns around stigma.[2,3]

Artificial intelligence (AI) applications have shown promise in the management of pandemics and have been widely used to assist the identification, classification, and diagnosis of medical images, such as in the case of the last PHEIC, COVID-19.[4] Others have also started to do so with Mpox.[5-9] In these studies, researchers created datasets through data augmentation, as well as using images of skin lesions caused by Mpox, chickenpox, measles, and normal controls collated from open-source data. Pre-trained, modified, and ensemble CNN were then used to classify Mpox from other classes, obtaining accuracies of between 83.0% to 99.0%.[10] A systematic review on the application of AI techniques for Mpox highlighted how AI has not only been used for the diagnostic testing or screening of Mpox, but also in the epidemiological modelling, drug and vaccine discovery, and media risk management in the context of Mpox.[10]

While our team had successfully deployed a tool that could screen for symptomatic sexually transmitted diseases (STDs) through visible dermatological clinical signs, we needed to rapidly adapt this tool to ensure that those who are at risk of acquiring Mpox had a similar option of assessing their risks for symptomatic Mpox without the fear of stigma, and guide them towards appropriate interventions where possible. We aimed to develop a robust digital screening test for symptomatic Mpox through AI and machine learning approaches.

We leveraged our existing AI symptom checker tool designed to detect STDs to do so. We sought to do so across five stages. First, we had initial discussions with medical, sociobehavioral, community and statistical experts were initiated to better understand the nature of the current Mpox PHEIC. Second, given the lack of available data in the global outbreak, through the proposed system, the team aimed to intensify stakeholder engagements under several key directions such as, (1) to amplify discussions on the clinical presentation and risk factors of Mpox in the ongoing pandemic, (2) enable community stakeholders to address the stigma attached to Mpox, and (3) healthcare institutions to discuss possibilities of using data to train the tool. Third, to consolidate images to adapt and train our existing AI tool for the screening of Mpox. Fourth, to validate the Mpox symptom checker tool. And finally, to continue refining and validating our Mpox symptom checker tool in community and healthcare settings.

**Methods**

*The HeHealth team*

A core interdisciplinary team drove the development and implementation of the AI tool. This team comprised a management consultant, product specialist, physicians, data scientists, a sociobehavioral researcher, a community representative, and other assistants who provided diverse insight and approaches to problem-solving and developing an effective clinical tool. Following the development of the tool and initial rollout, primary care providers were co-opted later through a wider network to further provide pragmatic validation of the tool in primary care settings.

*Initial development of the AI tool to screen for sexually transmitted diseases*

**Figure 1**

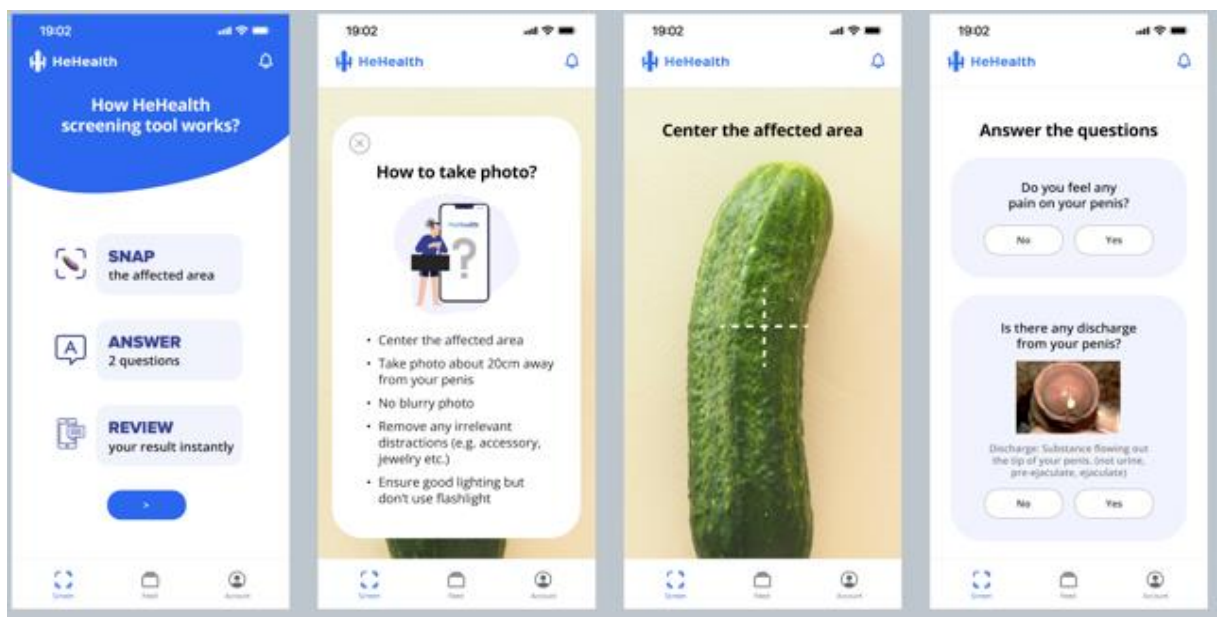

The team developed a smartphone app (HeHealth.ai) where app users can use their own smartphone cameras to take pictures of their own penises to screen for symptomatic STDs. **Figure 1** provides the symptom checker tool wireframe. The AI model was initially developed using 5000 cases and uses a modified convolutional neural network (CNN) to output prediction scores across visually diagnosable penis pathologies. The underlying AI predictions are conducted in two phases. In phase 1, given the input image of the affected area, a computer vision-based screening model outputs prediction scores across visually diagnosable penis pathologies. Note that the screening model was specially designed using modified convolutional neural network (CNN) and was initially trained using 5000 cases. In phase 2, users' answers to 2 simple questions (i.e., pain and discharge related) that are used to further fine tune the visual prediction results.

**Figure 2**

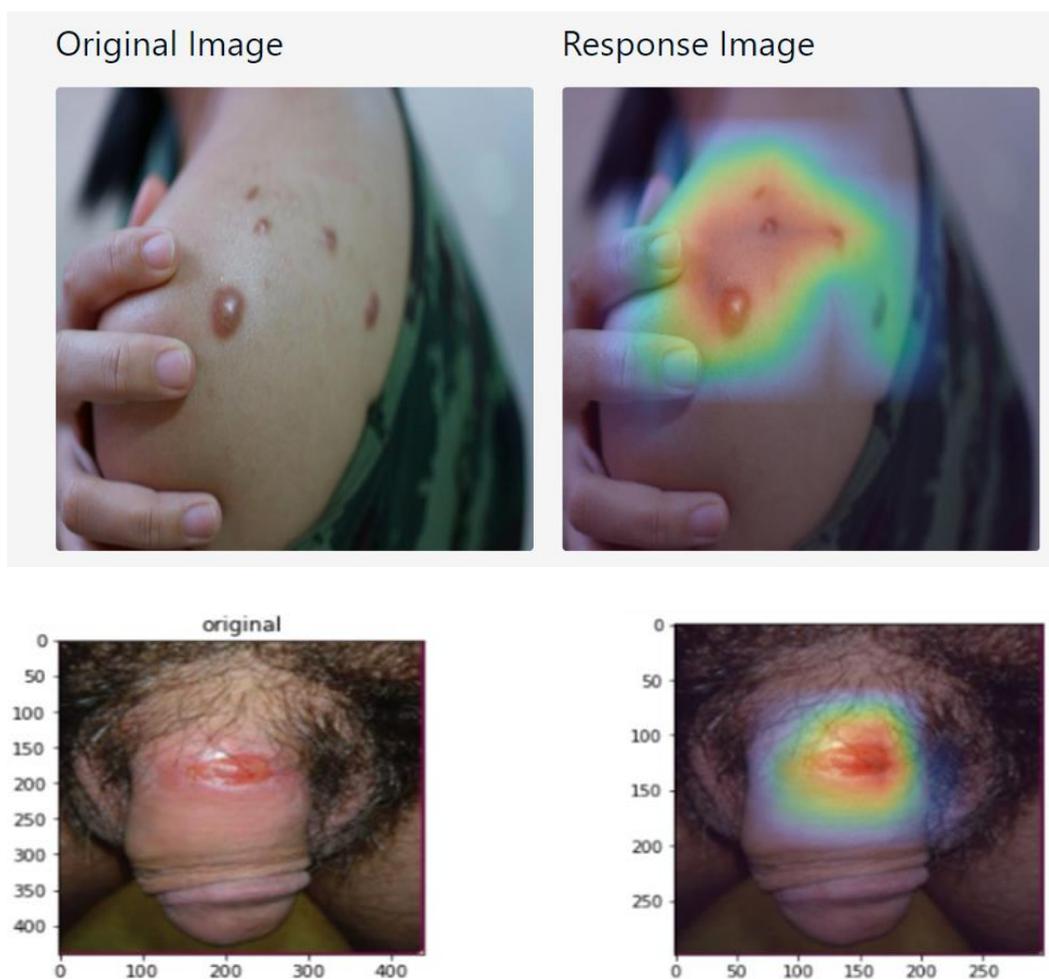

The interface then displayed the potential diagnoses that were output by the AI, sorted in the order of the AI's predicted likelihood scores. Artificial intelligence predictions with low likelihood scores (<40% or <0.4) are not shown and are removed, and the resulting display included a primary possible condition and secondary possible condition with confidence levels (in percentages) to avoid presenting extraneous information. In addition, a heatmap showing the detected pathological locations are displayed. **Figure 2** provides an example of the heatmap. We subsequently conducted validation of the first 1000 cases, during which each image is reviewed by a venereologist situated in Singapore, Sri Lanka, or the United Kingdom. Following the review, corresponding users who submitted the images were given feedback from the venereologists that could be accessed through the app by a user notification. Given that personal information is not collected through this process, such feedback can only be obtained through a unique username and password created by the user to log into the system.

*Adaptation of the AI platform for the Mpox symptom checker tool*

**Figure 3**

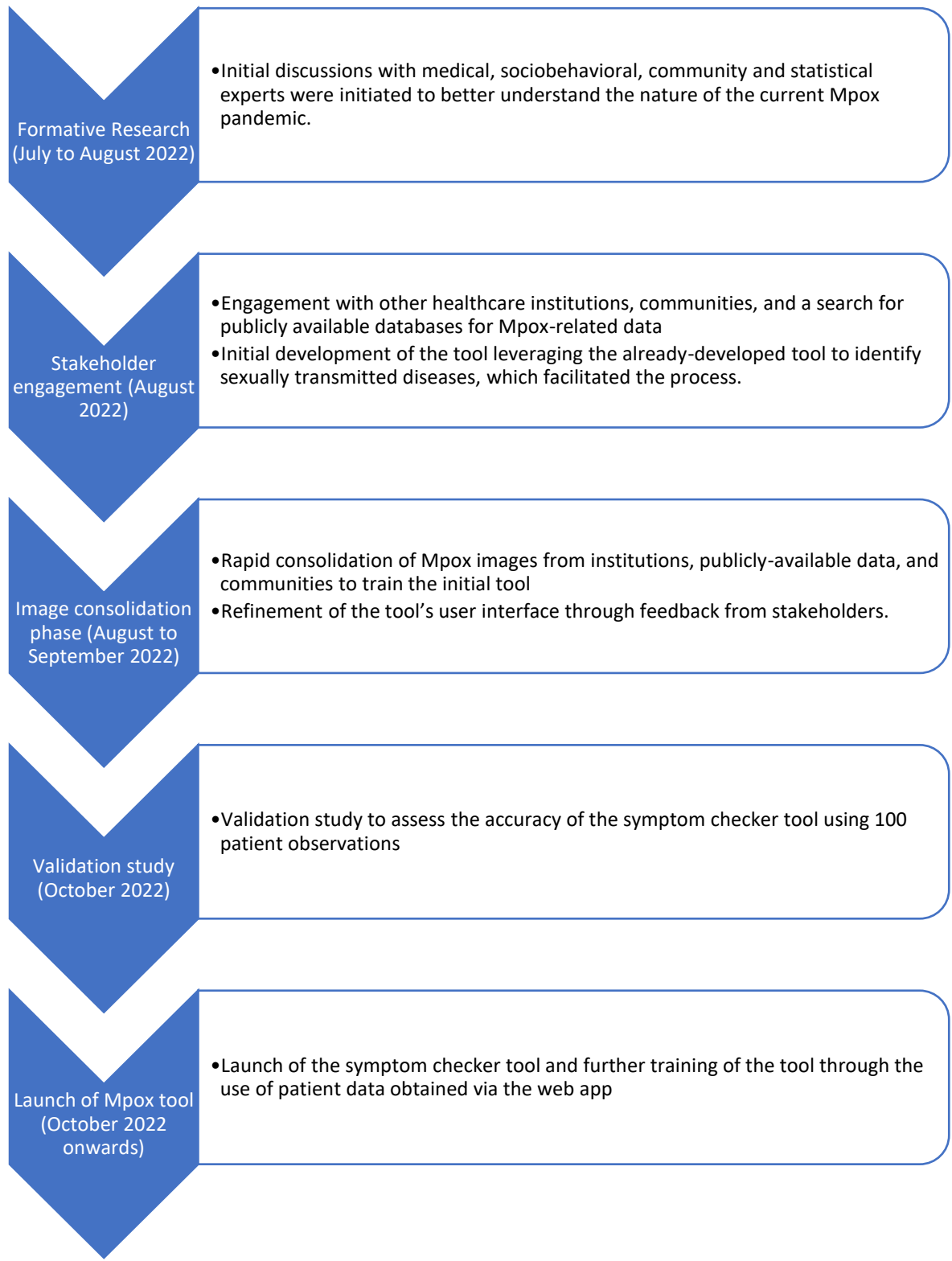

The overall timeline describing the development and subsequent refinement of the penis symptom checker for Mpox screening can be found in **Figure 3**. The first phase (July to August 2022) comprised of the formative phase where initial discussions with medical, sociobehavioral, community and statistical experts were initiated to better understand the nature of the current Mpox PHEIC. The second phase (August 2022) comprised of engagement with other healthcare institutions and a search for publicly available databases for Mpox-related data (e.g., patient characteristics, images etc.) and the initial development of the tool. This phase leveraged the already-developed tool to identify STDs, which facilitated the process. The third phase (August to September 2022) that was iterative in nature involved the rapid consolidation of Mpox images to train the initial tool, and refinement of the tool's user interface through feedback from stakeholders. The fourth phase (October 2022) involved a validation study to assess the accuracy of the symptom checker tool. The final phase (October 2022 onwards) involved the launch of the symptom checker tool and further training of the tool through the use of patient data obtained via the web app.

*Intensifying stakeholder engagement for tool development*

Given that user and patient trust were instrumental in designing a successful and effective app that is user-centred, we opted to ensure that the tool was backed by not just medical and public health experts, but also by community representatives and stakeholders. In general, we garnered deep insight from medical and public health practitioners on the key risk factors and drivers of the present Mpox global outbreak, leads for Mpox-related datasets and data from other researchers and industry partners, as well as insight on communications, language, and patient perspectives through community representatives.

*Risk scoring and validation of the Mpox symptom checker tool*

The HeHealth Symptom (H2S) score is a fundamental aspect of HeHealth's Mpox digital symptom checker tool. The tool was developed by physicians and scientists at HeHealth and is used to screen for symptomatic Mpox infection at primary care settings. The H2S score is based on an advanced algorithm developed by HeHealth and trained starting with a smaller dataset earlier in the global outbreak, which was then built upon to eventually be developed based on more than 10,000 data points including 1,000 Mpox positive rash images.

Mpox HeHealth Symptom (H2S) score calculator screenshot

**Figure 4.**

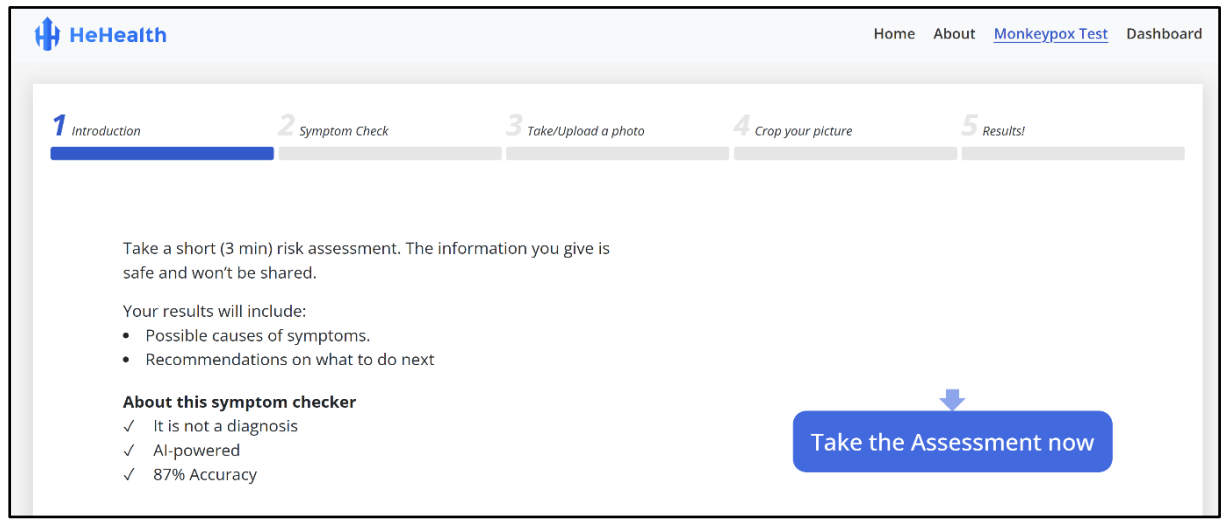

The algorithm was designed considering 3 key parameters (1) the World Health Organization and the United States Centres for Disease Control and Prevention's latest Mpox positive patient statistical data, (2) geographical prevalence data, (3) and HeHealth's patent-pending proprietary AI image analysis (computer vision) technology. Participants receive a percentage risk score where a score of 80% and higher means that they are highly likely to have a symptomatic Mpox infection, a score between 41-79% mean that further details are needed to differentiate such cases, and a score of 40% and below means that it is unlikely a Mpox infection.

*Ethics approval*

Ethics approval was obtained from the National University of Singapore Institutional Review Board (NUS-IRB-2023-983).

**Results**

HeHealth used 5000 images to train the initial AI tool for STDs identification. Of all the STDs, our tool can diagnosis for Syphilis, Herpes Simplex Virus (HSV), Human Papilloma Virus (HPV) / Genital Viral Warts with accuracy of 86%, 93%, and 96%, respectively. From June 2022 to October 2022, a total of about 22,000 users had downloaded the HeHealth app, and about 21,000 images have been analysed using HeHealth AI technology. In the same period, we had a total of 1000 number of Mpox-related images that have been used to train the Mpox symptom checker tool. Our digital symptom checker tool shows accuracy of 87% to rule in Mpox and 90% accuracy to rule out the symptomatic infection.

After the initial training, the algorithm had undergone testing and validation using 100 patient observations (60 Mpox positive patients and 40 Mpox negative patients with similar rashes/conditions). Based on this dataset, HeHealth's current Mpox digital symptom checker test (V1.0) was able to diagnose Mpox (rule-in) with an accuracy of 87% and exclude the non-Mpox infection (rule-out) with an accuracy of 90%. HeHealth's Mpox digital symptom checker test is currently undergoing further validation through community and clinical partnerships.

**Discussion**

Throughout the process of developing the innovation and field testing it, we noted several lessons learnt that arose due to several limitations or hurdles and conclude this article with recommendations for other innovators who are keen to work on similar projects. The first hurdle was about data privacy and security for app users. In our formative research conducted among potential users of the app, many of them raised issues of data privacy and security with us. This was a hurdle that the team had to overcome prior to the launch of the original STDs symptom checker tool, which would have had implications for patient confidentiality and trust in the use of the Mpox symptom checker tool. Users generally feel safe using the HeHealth tool as no personal identifiers are collected throughout the user journey. Any collected data including images are deidentified in ways that cannot be traced back to the users. All processes are conducted through the Health Insurance Portability and Accountability Act of 1996 (HIPAA)-compliant processes and platforms.

The second hurdle came about due to an initial lack of data to train the AI model. Initially, health system and professionals were approached for the images. But with less success due to regulatory restrictions such as patient confidentiality. Furthermore, the lack of a central coordinating body for data consolidation (e.g., images of disease state) served as a key barrier to training the AI model early on. The third hurdle was due to technical challenges. Given that the current outbreak has been characterised by novel clinical presentation and patient characteristics,[11] our team needed to be in constant dialogue with expert stakeholders, as well as to constantly monitor and review the scientific evidence surrounding the current outbreak of Mpox. Furthermore, unlike STDs, we had to train out AI tool to also review ulcers and lesions that appeared on other parts of the body that were not limited to the anogenital region.

The fourth and final hurdle arose due to concerns around the generalizability of our data. One concern arose during our earlier AI tool development was ensuring that our tool was inclusive of all skin colours. We ensured this was done through several means. First, we tried to build a dataset that represented multiple skin colours from a global population. At the same time, our core technology was built in a way skin results will not be affected by skin color by training the AI to focus on colour-neutral skin abnormalities. Another concern was that pictures of visible skin lesions meant that symptomatic STDs that present in other ways including other dermatological conditions or genitourinary symptoms may get left out. Nevertheless, we used a combination of questionnaires they addressed issues like the presence of discharge or pain to accompany the visual analysis of skin lesions, which may also help address and narrow down the various types of potential STDs that an individual may be experiencing.

**Conclusions**

Throughout the process of developing the original STDs symptom checker tool and the Mpox symptom checker tool, we took away several key lessons that we believe would be useful for researchers and other practitioners interested in embarking on similar innovations in healthcare. First, to engage a wide range of stakeholders early. In the context of an ongoing global pandemic, and with rapidly evolving data and clinical insights, we found that engaging not only the medical science experts, but also community representatives were important. Collaborators shared similar goals in helping to alleviate the impact of Mpox around the world and were willing to work together in achieving similar outcomes. Second, to have a multidisciplinary team. Our multidisciplinary team provided not just expertise in executing various aspects of work in the rollout of the symptom checker tool, but also provided strategic directions that led towards the prioritization of key development milestones, including stakeholder engagement, building the AI tool, the validation study, and eventual rollout of the tool.

Third, to prioritize pragmatism over research 'elegance'. Initial efforts to validate the initial tool to screen STDs had met several barriers – this included regulatory barriers in Singapore due to being classified as a medical device, as well as the requirement of the local university ethics board in obtaining approvals from the relevant medical devices authorities. While it is important that formal research be conducted to validate the accuracy of the tool, this was done at a later stage prior to rollout. Nevertheless, all processes were HIPAA-compliant and adhered to best clinical practices. And finally, that big data comprises small data. Data for our AI tool were consolidated incrementally, which led to the eventual consolidation of about 10,000 Mpox and STD-related images. This was, to our knowledge, one of the largest and most comprehensive datasets globally. Instead of waiting on a wider body of data, researchers should consider engaging potential partners in the sharing of data in secure ways, so that an active rather than passive approach is undertaken.

**Declaration of Funding:** This study was funded by HeHealth.ai.

**Data Availability Statement**: The data that support this study cannot be publicly shared due to ethical or privacy reasons and may be shared upon reasonable request to the corresponding author if appropriate.

**Table 1. List of stakeholders engaged to develop the Mpox tool**

| Types of stakeholders | Insights |
|---|---|
| World Health Organization representative | • Data and statistics regarding the risk prevalence |
| Centers for Disease Control | • Data and stats regarding the risk prevalence |
| Industry Partners | • Research insights and validation |
| Community Representatives | • Communications strategy for Mpox risk and symptom checker tool<br>• Sources of stigma and patient care pathway for Mpox<br>• User interface and language used for key populations at risk of Mpox |
| General public | • Mpox images |